\documentclass[nohyperref]{article}

\usepackage{microtype}
\usepackage{graphicx}
\usepackage{caption}
\usepackage{subcaption}
\usepackage{booktabs} 

\usepackage{hyperref}

\usepackage[accepted]{ai4science} 

\usepackage{amsmath}
\usepackage{amssymb}
\usepackage{mathtools}
\usepackage{amsthm}
\usepackage{glossaries}

\usepackage[capitalize,noabbrev]{cleveref}

\theoremstyle{plain}

\theoremstyle{definition}

\theoremstyle{remark}

\usepackage{todonotes}
\usepackage{comment}
\usepackage{todonotes}
\usepackage[section]{placeins}

\glsdisablehyper
\newacronym{APL}{APL}{%
  the Johns Hopkins University Applied Physics Laboratory%
}

\newacronym{CV}{CV}{computer vision}

\newacronym{DFT}{DFT}{density functional theory}

\newacronym[shortplural={GNs}]{GN}{GN}{graph network}

\newacronym{JHU}{JHU}{%
  Johns Hopkins University%
}

\newacronym{MAE}{MAE}{mean absolute error}

\newacronym{ML}{ML}{machine learning}

\newacronym{MTL}{MTL}{multi-task learning}

\newacronym{MP}{MP}{Materials Project}

\newacronym{NLP}{NLP}{natural language processing}

\newacronym[
    shortplural={PINNs}
]{PINN}{PINN}{physics-informed neural network}
\newacronym{PDE}{PDE}{partial differential equation}

\icmltitlerunning{Curvature-informed multi-task learning for graph networks}

\begin{document}

\twocolumn[
\icmltitle{Curvature-informed multi-task learning for graph networks}



\icmlsetsymbol{equal}{*}

\begin{icmlauthorlist}
\icmlauthor{Alexander New}{apl}
\icmlauthor{Michael J. Pekala}{apl}
\icmlauthor{Nam Q. Le}{apl}
\icmlauthor{Janna Domenico}{apl}
\icmlauthor{Christine D. Piatko}{apl}
\icmlauthor{Christopher D. Stiles}{apl}
\end{icmlauthorlist}

\icmlaffiliation{apl}{Research and Exploratory Development Department, Johns Hopkins University Applied Physics Laboratory}

\icmlcorrespondingauthor{Alexander New}{alex.new@jhuapl.edu}

\icmlkeywords{Machine Learning, ICML}

\vskip 0.3in
]



\printAffiliationsAndNotice{}  

\begin{abstract}
Properties of interest for crystals and molecules, such as band gap, elasticity, and solubility, 
are generally related to each other: they are governed by the same underlying laws of physics.
However, when state-of-the-art graph neural networks attempt to predict multiple properties simultaneously (the \gls{MTL} setting), they frequently underperform a suite of single property predictors.  This suggests graph networks may not be fully leveraging these underlying similarities.
Here we investigate a potential explanation for this phenomenon – the curvature of each property's loss surface significantly varies, leading to inefficient learning. This difference in curvature can be assessed by looking at spectral properties of the Hessians of each property's loss function, which is done in a matrix-free manner via randomized numerical linear algebra.
We evaluate our hypothesis on two benchmark datasets (\gls{MP} and QM8) and consider how these findings can inform the training of novel multi-task learning models.
\end{abstract}

\glsresetall

\section{Introduction}

\begin{figure}
    \centering
    \includegraphics[width=0.85\linewidth]{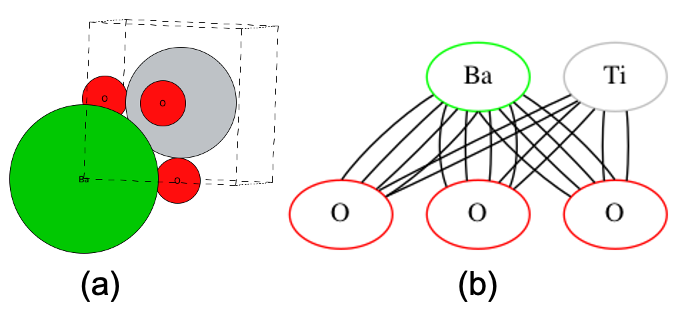}
    \caption{\textbf{Example of converting a periodic crystal lattice structure ($\text{Ba}\,\text{Ti}\,\text{O}_3$) to a multigraph}\\ (a): Crystal lattice structure, created using \texttt{ase}\footnote{\url{https://wiki.fysik.dtu.dk/ase/}}.
    \\ (b): The corresponding multigraph representation of the crystal that is ingested by a graph network.}
    \label{fig:example_materials}
\end{figure}

\begin{figure}
    \centering
    \includegraphics[width=0.9\linewidth]{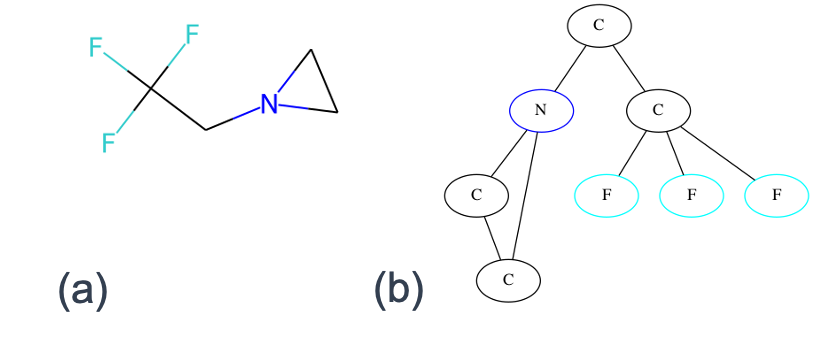}
    \caption{\textbf{Example of converting a 2D molecule structure ($\text{C}_4\,\text{H}_6\,\text{F}_3\,\text{N}$) to a graph}\\(a) Molecular structure as represented by \texttt{rdkit}\\(b) The corresponding graph}
    \label{fig:chem_sample}
\end{figure}

\Glspl{GN}~\cite{Battaglia2018graphnetworks} are considered state-of-the-art \gls{ML} methods for many scientific problems, including property prediction of both inorganic crystalline materials~\cite{Xie2018CGCNN,Park2020icgcnn} (hereafter ``crystals'', e.g.,~\cref{fig:example_materials}) and small organic molecules~\cite{Gilmer2017Neural,Gasteiger2020dimenet} (hereafter ``molecules'', e.g.,~\cref{fig:chem_sample}); the same \gls{GN} architectures have shown success in both domains (e.g.,~\cite{chen2019megnet}, who suggest the ``crystal'' vs. ``molecule'' terminology and use ``material'' to refer to both). A common use case might involve training a model from materials with known properties and then screening a separate dataset lacking those properties to obtain potential candidates to further investigate. Typically, property-prediction tasks are formulated as single-target regression problems, where the goal is to predict a scalar property like band gap, formation stability, elasticity, or solubility.
However, practical materials design requires optimizing multiple properties of interest. For example, when designing a cell phone screen, a material with both high optical transparency and hardness would be desired. 

Predicting a single property can be posed as a single task for an \gls{ML} model; using a single 
\gls{ML} model to predict multiple properties is thus a type of \gls{MTL}. The most common family of \gls{MTL} approaches is hard parameter sharing, in which a single representation space is shared across multiple tasks of interest, and task-specific networks map from that space to output predictions~\cite{Caruana1997mtl}.

However, the latest research in novel \gls{MTL} techniques largely has been focused on classification problems in the \gls{NLP} and \gls{CV} domains -- for example, the $\mathrm{MultiMNIST}$ problem~\cite{Sener2018Multiobjective}, where a model must identify two separate digits in the same picture. With some recent exceptions -- e.g.,~\cite{Yang2021multitask,Tan2021multitask,Kong2021multitask} -- there has been little work exploring the application of \gls{MTL} to multi-property \gls{GN} regression problems.

Some existing papers have used hard parameter sharing for material property prediction, in which a single model predicts a set of properties. However, results suggest that this approach has degraded performance compared to a set of single-property models -- see, e.g.,~\cite{Gasteiger2020dimenet}, which predicts twelve quantum mechanical properties of molecules and finds that a single multi-output model performs worse than a set of single-output models.

Works have explored reasons for this: \cite{Yu2020Surgery} identifies a ``tragic triad'' of reasons that \gls{MTL} models might perform worse compared to single-task ones, one of which is that multi-task loss functions can have high curvature, as characterized by the Hessian of the loss. However,~\cite{Yu2020Surgery} rely on first-order approximations to curvature and do not directly incorporate second-order information.

Concurrently, several works have applied techniques from randomized numerical linear algebra to directly probe the Hessians of large \gls{ML} models~\cite{Sagun2017hessian,Alain2019negativeeigenvalues,Yao2019pyhessian,Ghorbani2019density,Papyan2020spectra}. These rely (1) on the fact that the product of a model's Hessian and another vector (a Hessian-vector product) may be efficiently obtained, without explicitly calculating the full Hessian, and (2) that given a matrix-free Hessian-vector product function, several key spectral properties of the Hessian, including its eigenvalues and trace, may be estimated. However, these papers focus only on standard image classification problems, and do not consider the \gls{MTL} setting, \glspl{GN}, or regression problems.

In this paper, we formulate the application of curvature-informed techniques to \gls{MTL} models (\cref{sec:curvature}), describe the application of graph networks to the problem of multi-property prediction in the domains of materials science and chemistry (\cref{sec:prediction} and \cref{sec:data}), and analyze training dynamics of multi-property prediction models using methods based on assessing local curvature of loss function surfaces (\cref{sec:results}). Our results suggest how novel multi-property prediction models might inherently account for differences in curvature to enhance learning efficiency. 


\section{Approach}\label{sec:approach}

Let $\mathcal{G}$ be a space of directed multigraphs corresponding to crystal or molecule structures. A directed 
graph $g = (\mathcal{V}, \mathcal{E})$ consists of a set of nodes $\mathcal{V} = \{v\}$ and edges $\mathcal{E} = \{(v,v',k)\}$. Each node $v$ has a node
embedding $x_v \in \mathbb{R}^V$, and each edge $(v,v',k)$ has an edge embedding $e_{v,v',k} \in \mathbb{R}^E$.  A pair of nodes $v$ and $v'$ may have multiple edges connecting them, and these multi-edges are indexed by $k = 1,\hdots,K_{v,v'}$. Furthermore, a graph $g$ has $T$ properties $y = (y_1,\hdots,y_T) \in \mathcal{Y}$, here assumed to all be scalars. 

We require a model that can predict all targets $y$ for a given $g$. This entails minimizing a loss with respect to a set of neural network weights $\{\theta_{sh},\theta_1,\hdots,\theta_T\}$, where $\theta_t$ is used only in predicting property $t$, and $\theta_{sh}$ is shared across predictions. The form of this problem and its neural network is discussed in~\cref{sec:prediction}. First we consider the loss's curvature.

\subsection{Curvature assessment techniques}\label{sec:curvature}


We view the shared parameters $\theta_{sh}$ of a graph network as a flattened vector in $\mathbb{R}^P$; let $L_t:\mathbb{R}^P \to \mathbb{R}$ be a property-specific loss function of $\theta_{sh}$, where the property-specific parameters $\theta_1,\hdots,\theta_T$ are held constant. Typically, machine learning models consider only first-order derivative information of $L_t$ when training neural networks. However, second-order information is also useful in optimization problems -- consider how a quasi-Newton method like limited-memory L-BFGS~\cite{Liu1989lbfgs} is more efficient than a first order method like gradient descent. 
This suggests that additional insight into the training problems of \gls{ML} models can be found in using second-order information.

The curvature of a function $L_t$ is characterized by its Hessian $H_t = \nabla_{\theta_{sh}}^2 L_t \in \mathbb{R}^{P \times P}$ for a weight vector $\theta_{sh} \in \mathbb{R}^P$. For typical deep neural networks, $P \approx 10^6$, and calculating $H$ directly is computationally infeasible due to storage contraints ($H_t$ will have $10^{12}$ entries). However, the product of $H_t$ with a vector $v$ is computable in approximately the same time-complexity as calculating the gradient of $L_t$ with respect to $\theta$~\cite{Pearlmutter1994hessian}.\footnote{In \texttt{jax}, this is implemented by composing a forward-mode Jacobian-vector product \texttt{jvp} function with a reverse-mode gradient function \texttt{grad}.} This enables calculation of properties of $H_t$ that can be obtained from observing its action on a chosen vector $v$~\cite{Yao2019pyhessian,Ghorbani2019density}.
In particular, the trace of $H_t$, and an approximate probability density function of its eigenvalues can be efficiently estimated.

The trace of $H_t$ is both the sum of its eigenvalues and $L_t$'s Laplacian (i.e., $\mathrm{tr}\, H_t = \sum_p \partial_{\theta_{sh,p}\theta_{sh,p}} L_t(\theta)$). Thus, the trace of $H_t$ can be viewed as a measurement for the complexity of its curvature~\cite{Yao2019pyhessian}. The sign of the trace also indicates the sign of $L_t$'s local curvature. The trace of $H_t$ may be estimated with the expectation 
$$\mathrm{tr}\, H_t = \mathbb{E}_{v\sim\mathbb{P}_R} [v^T H_t(\theta) v],$$
where $\mathbb{P}_R$ is the distribution of random variables in $\mathbb{R}^P$ with $iid$ Rademacher components~\cite{Avron2011trace}. 

Having access to all of $H_t$'s eigenvalues gives us a full picture of $L_t$'s curvature. These eigenvalues are often represented as a function $\psi$ called the spectral density or density of states that is given by $\psi(t) = \frac{1}{P} \sum_p \delta(t - \lambda_p)$, where $\lambda_1 \leq \cdots \leq \lambda_P$ are the eigenvalues of $H_t$, and $\delta$ is the Dirac distribution~\cite{Lin2016lanczos}. However, the weight vector $\theta_{sh}$ is too high-dimensional for all of $H_t$'s eigenvalues to be directly calculated in practice. Methods like power iteration can be used to estimate the large-magnitude eigenvalues of a Hessian~\cite{Yao2019pyhessian}, but they scale poorly as its dimensionality increases.

A solution is to first use the Lanczos algorithm to estimate $P' \ll P$ eigenvalues of $H_t$ and then smooth these estimated values into a continuous approximation to the discrete density $\psi$ by convolving the values with a Gaussian kernel~\cite{Golub1969gauss,Ghorbani2019density}.\footnote{The Lanczos algorithm accurately estimates the Hessian's most positive and most negative eigenvalues, and the convolution approximates the distribution of eigenvalues between these extrema.} This yields a continuous function $\hat{\psi}(t)$ that approximates $\psi$. See~\cref{sec:extra-lanczos} for an example of estimating $\hat{\psi}$ for a simplified loss function.


Interpreting estimated spectral densities remains a challenging task:
\noindent\cite{Ghorbani2019density} and~\cite{Yao2019pyhessian} claim that a spectral density should be ``smooth'' (i.e., concentrated within a dense region) and consider the effect of different network architectures on density smoothness. \cite{Alain2019negativeeigenvalues} argues that having a large concentration of negative eigenvalues can lead to inefficient training, because most optimizers fail to leverage this local negative curvature. \cite{Papyan2020spectra} relate the outlier structure of a Hessian's eigenvalues to the number of classes in a dataset. 

Compared to \gls{CV} problems, graph-structured datasets have a number of interesting attributes. For example, a graph input has a variable number of node and edge features, which complicates the~\gls{GN} learning problem. Furthermore, in contrast to the use of very deep networks in \gls{CV} problems, increasing depth in graph networks has often not been found to be effective (see, e.g., discussion in~\cite{Godwin2022simple}). Hessian-based information has the potential to inform some of these comparisons.

Note that the estimations of both the trace and the spectral densities are based on random quantities. For trace, multiple vectors $v$ must be sampled and the quantity $v^T H_t(\theta) v$ computed, and then convergence in mean can be assessed. In practice, we find that approximately $500$ samples are sufficient. For spectral density, each Lanczos iterate is initialized as a random Gaussian vector. We adapt an implementation of the Lanczos algorithm and smoothing process developed by~\cite{Ghorbani2019density}, in which we reorthogonalize the Lanczos iterates during each step of estimation to promote numerical stability. We follow~\cite{Yao2019pyhessian} and use $P' = 100$, which is more than the $P'=90$ that~\cite{Ghorbani2019density} validate by comparing to an exact calculation of every eigenvalue of a smaller network.

\subsection{Graph networks for property prediction}\label{sec:prediction}


Let $\hat{y}:\mathcal{G} \to \mathcal{Y}$ be a neural network parameterized by $\theta$. We follow~\cite{Sener2018Multiobjective} in splitting $\theta$ into a set of shared weights $\theta_{sh}$, and a set of property-specific weights $\theta_1,\hdots,\theta_T$.
Then the predicted $t$\textsuperscript{th} property is
$$\hat{y}_t(g; \theta_{sh}, \theta_t) = f_t(z(g; \theta_{sh}); \theta_t),$$

where $z:\mathcal{G} \to \mathbb{R}^D$ is a task-independent \gls{GN}~\cite{Battaglia2018graphnetworks}, and each $f_t: \mathbb{R}^D \to \mathbb{R}$ is a task-specific feedforward network.

We briefly outline the functionality of our graph network $z$, which maps an input graph $g$ to a graph-level representation vector $u^M$. 
First, the multigraph $g$ is given a global feature $u^0 \in \mathbb{R}^U$ that is initialized as a vector of ones, and, similarly to~\cite{chen2019megnet}, the node features $x_v$ and edge features $e_{v,v',k}$ are linearly projected: $x^0_v = W_V\,x_v$ and $e_{v,v',k}^0 = W_E\,e_{v,v',k}$, where $W_V$ and $W_E$ are matrices learned during training. Then information is propagated across the multigraph during $m=1,\hdots,M$ message-passing steps. 

The $m$\textsuperscript{th} message-passing step proceeds as follows: First, the edge features are updated:
$$e^m_{v,v',k} = \phi^E(\mathrm{cat}(e_{v,v',k}^{m-1}, x_v^{m-1}, x_{v'}^{m-1}, u^{m-1})) + e^{m-1}_{v,v',k},$$
where $\phi^E$ is a feed-forward neural network, and $\mathrm{cat}$ is the concatenation operator. Then node updates are collected. For every node $v$, first edge updates are calculated:
\begin{eqnarray*}
h_{v,out}^m &=& \sum_{(v',k) : (v,v',k) \in \mathcal{E}} e^m_{v,v',k}\\
h_{v,in}^m &=& \sum_{(v',k): (v',v,k) \in \mathcal{E}} e^m_{v',v,k},
\end{eqnarray*}
where the first summation is over the tuples $(v',k)$ such that $(v,v',k)$ is an edge for $g$, and the second summation is defined similarly. With these updates, the new node representation for $v$ is given by
$$x^m_v = \phi^V(\mathrm{cat}(h_{v,out}^m, h_{v,in}^m, x^{m-1}_v, u^{m-1})) + x^{m-1}_v,$$
where $\phi^V$ is a feed-forward neural network. Finally, the global feature is updated:
$$u^m = \phi^U\left(\mathrm{cat}\left(\sum_{v,v',k} e^m_{v,v',k}, \sum_v x^m_v, u^{m-1}\right)\right) + u^{m-1},$$
where $\phi^U$ is a feed-forward neural network, and each feature vector $x^m_v, e^m_{v,v',k}, u^m$ is processed with a layer normalization~\cite{Ba2016LayerNorm} layer. 
After $M$ steps, the final global state $u^M$ is fed into each property-specific network $f_t$. 


Our graph networks are implemented in the \texttt{jraph}\footnote{\url{https://github.com/deepmind/jraph}} framework that is built on \texttt{jax}\footnote{\url{https://github.com/google/jax}}. We use \texttt{flax}\footnote{\url{https://github.com/google/flax}} to implement component neural networks $\phi^V$, $\phi^E$, and $\phi^U$. Because we seek to evaluate second-order derivative information, we ensure that our neural networks $\phi^V,\phi^E,\phi^U$ are smooth functions with respect to their weights by using $\tanh$ activation functions. Further details on specific hyperparameters used to instantiate models are given in~\cref{app:model}. 

We impose on each property a loss function $L_t$ (here assumed to be mean squared loss for all $t$), and we collect a training dataset of multigraph data points $\mathcal{D} = \{(g^n, y^n)\}_{n=1}^N$. 
Then we solve the minimization problem
$$\min_\theta L(\theta) =  \sum_t L_t(\theta_{sh}, \theta_t),$$
where 
$$L_t(\theta_{sh}, \theta_t) = \sum_n |\hat{y}_n(g^n; \theta_{sh}, \theta_t) - y^n_t|^2$$
are task-specific losses and
$\theta = \{\theta_{sh}, \theta_1,\hdots,\theta_T\}$.

%
We solve this minimization with stochastic gradient descent using the \texttt{optax}\footnote{\url{https://github.com/deepmind/optax}} implementation of AdamW~\cite{Loshchilov2018adamw} to choose stepsizes; we handle the potential difference in scales across properties by normalizing all targets prior to training and set $\mu_t = 1$ for all $t$. Further details on the specifics of model training are given in~\cref{app:training}.

\subsection{Data sources}\label{sec:data}


We evaluate our graph network curvature assessment techniques on datasets from two scientific domains: materials science (crystals) and chemistry (molecules). For crystals, our data featurization follows~\cite{Park2020icgcnn}; for molecules, it follows~\cite{Gilmer2017Neural}. To reduce the complexity of our considered problem space, we choose to use simple featurizations (one-hot encoding of atom-types as node featurizations), but our methods are compatible with other featurizations (e.g., hand-crafted node descriptors like those provided by \texttt{rdkit}). 
Further details about the data used are available in~\cref{app:data}.

For materials science, we collect data from \gls{MP}~\cite{Jain2013MaterialsProject}, which contains results of \gls{DFT} calculations for different crystals. crystal records contain compositional and structural information, as well as some related properties. Each crystal's structure information is captured in a Crystallographic Information File (CIF), which we convert into a multigraph using the \texttt{VoronoiNN} function from \texttt{pymatgen}\footnote{\url{https://pymatgen.org}},
following the approach of~\cite{Park2020icgcnn}. As crystals are periodic structures, this process yields multiple edges between many given pairs of nodes~\cite{Xie2018CGCNN,Park2020icgcnn,Sanyal2018MTCGCNN} (e.g.,~\cref{fig:example_materials}). As targets, we use several assessments of a crystal's elastic properties: the Voigt-Reuss-Hill~\cite{Hill1952vrh} calculation for shear ($G_{VRH}$) and bulk ($K_{VRH}$) modulus, both measured in units of GPa, as well as the isotropic Poisson ratio $\mu$, a dimensionless quantity. 

These properties are inherently coupled -- $\mu$ can be calculated as a function of $G_{VRH}$ and $K_{VRH}$. Thus, an \gls{ML} model that predicts all three serves as a test for how well it can learn underlying physical relationships. A one-hot encoding of atom-type is used to obtain initial node features $x_v$, and four bond-related properties calculated by \texttt{pymatgen} are used as edge features, resulting in a dataset of 10,500 crystal structures with known elastic properties.



For chemistry, we use the QM8~\cite{Ramakrishnan2015QM8} dataset of organic molecules. We use the \texttt{rdkit} package\footnote{\url{https://github.com/rdkit/rdkit}} to convert SMILES strings~\cite{Weininger1988smiles} into molecular structure graphs (see, e.g.,~\cref{fig:chem_sample}).
As targets, we use the first and second transition energies, $E_1$ and $E_2$, and the first and second oscillator strengths $f_1$ and $f_2$. QM8 contains several versions of transition energies and oscillator strengths that have been calculated via different levels of \gls{DFT}. We use the predictions of the approximate coupled-cluster (CC2)~\cite{Haettig2000cc2} method as our targets, as these are treated as the ``ground truth'' in~\cite{Ramakrishnan2015QM8}. A one-hot encoding of atom-type is used to obtain initial node features $x_v$, and a one-hot encoding of bond-type is used to obtain initial edge features $e_{v,v',k}$.

Note that, in~\cref{sec:prediction}, we describe our input data points $g$ as directed multigraphs. However, for our actual data points, the edges are not inherently directed -- bonds are symmetric. Thus, we duplicate each edge feature (i.e., $e_{v,v',k} = e_{v',v,k}$ for all edges) during data preprocessing.

\newpage


\section{Results}\label{sec:results}

\begin{figure}[t]
    \centering
    \includegraphics[width=\linewidth]{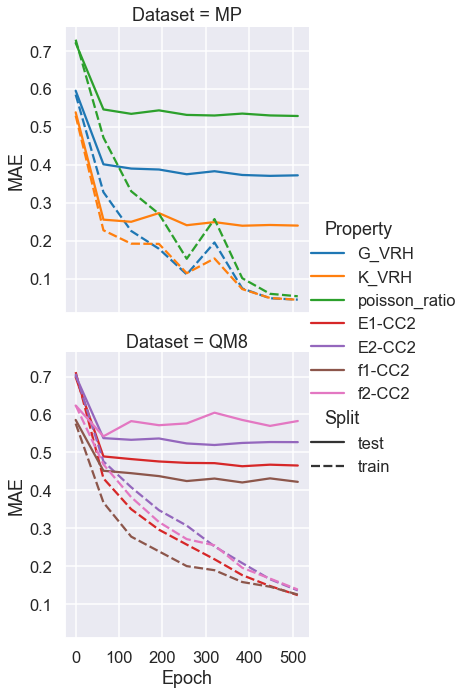}
    \caption{Standardized \gls{MAE} values for each property across train and test splits, for the \gls{MP} and QM8 prediction tasks. Despite the high variability in curvature across training (\cref{fig:trace_figs}), training error reduces relatively smoothly, with occasional regressions in performance for \gls{MP}. Although the training error for each property converges to rough equality, significant inter-property variety is observed for test set error.}
    \label{fig:train_errors}
\end{figure}

We demonstrate the application of our curvature-assessment techniques (\cref{sec:curvature}) on trained \glspl{GN} (\cref{sec:prediction}) in two application domains: materials science and chemistry (\cref{sec:data}). Although here we focus on \glspl{GN}~\cite{Battaglia2018graphnetworks}, our assessment techniques are applicable to the broad family of graph neural networks used in property prediction tasks.

\cref{fig:train_errors} shows training and test set errors for each dataset. These results do not necessarily align with domain intuition, suggesting that the models do not leverage scientific knowledge in their learned representations. For example, for \gls{MP}, the test set error for Poisson ratio is higher than that of $G_{VRH}$ or $K_{VRH}$, despite the fact that Poisson ratio can be calculated as a function of $G_{VRH}$ and $K_{VRH}$. This is interesting and suggests that the \glspl{GN} are not fully leveraging known scientific knowledge. For QM8, the oscillator strengths $f_1$ and $f_2$ have the lowest and highest, respectively, test set errors.

\begin{figure*}[htbp]
    \centering
    \begin{subfigure}{0.75\textwidth}
         \centering
         \includegraphics[width=\textwidth]{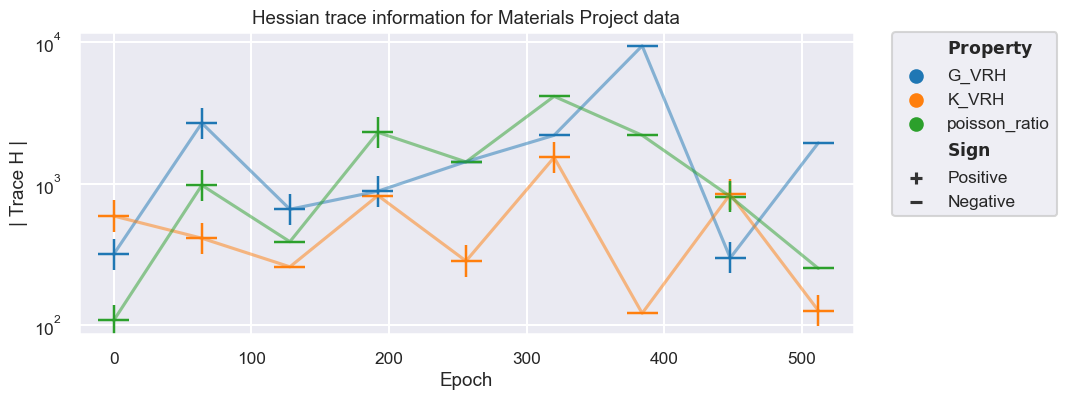}
         \caption{\gls{MP}}
         \label{fig:mp_trace}
     \end{subfigure}
     \hfill
     \begin{subfigure}{0.75\textwidth}
         \centering
         \includegraphics[width=\textwidth]{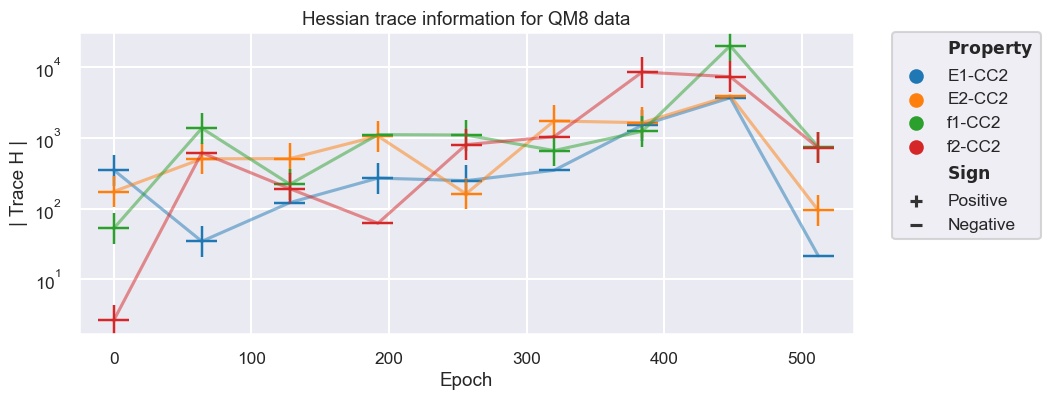}
         \caption{QM8}
         \label{fig:qm8_trace}
     \end{subfigure}
    \caption{We track property-specific Hessian traces $\mathrm{tr}\,H_t$ while training \glspl{GN} for the \gls{MP} and QM8 prediction problems. The loss surface of the \gls{MTL} problem for \glspl{GN} appears to be complicated and varying across properties. Trace signs (indicating the sign of the curvature) and trace magnitudes vary across training and across properties, the latter by orders of magnitude. In prior work that looked at traces of Hessians~\cite{Yao2019pyhessian} in \gls{CV} tasks, traces were found to be less variable.}
    \label{fig:trace_figs}
\end{figure*}

Next we summarize the curvature of the loss functions using the trace of each task's loss's Hessian. In~\cref{fig:trace_figs}, we show that, despite the general decreasing training error, the curvature of each property's loss surface, as measured by the trace of its Hessian, is highly variable. In particular, both datasets show high heterogeneity across properties and across training epochs in their estimated traces. Our observations here do not align with prior work that has analyzed Hessian traces for \gls{CV} problems~\cite{Yao2019pyhessian}. In those works, traces increased monotonically during training and were consistently positive. Here, our estimated traces often flip between being positive and negative.

\begin{figure*}
    \centering
    \includegraphics[width=0.9\linewidth]{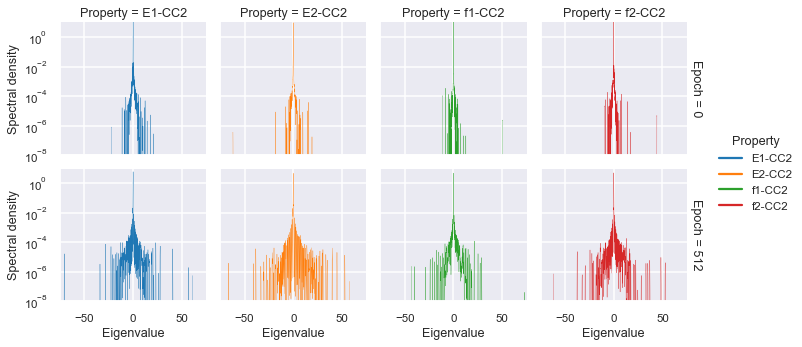}
    \caption{Snapshots of the spectral densities of property-specific Hessian losses for the QM8 dataset. All properties start with comparable densities closely concentrated near 0; but, as training finishes, the densities spread out as the loss surface increases in complexity. Each property has a different spread of eigenvalues, suggesting that the loss surfaces do vary in curvature by property. Note that this plot trims the range of the $x$-axis to remove outlier eigenvalues. The full spectra are displayed in~\cref{fig:full_spectra} in~\cref{sec:supp-figs}. For these plots, the $x$-axis gives the range of eigenvalues for the loss function, and the $y$-axis gives the density $\hat{\psi}(t)$ of eigenvalues concentrated at that point $t$ (as described in~\ref{sec:curvature}).}
    \label{fig:qm8_spectra_zoom}
\end{figure*}

The Hessian trace is a high-level summary of the curvature of a loss surface; for a more granular examination, we estimate the spectral densities of each property's Hessian. In~\cref{fig:mp_spectra_full} in~\ref{sec:supp-figs}, we show that the spectral densities are similarly variable and often feature the presence of outliers that briefly occur during training. These outliers vary across properties. For the QM8 dataset, we zoom in on the range of eigenvalues where most density is concentrated in~\cref{fig:qm8_spectra_zoom}. Similar to~\cite{Yao2019pyhessian,Ghorbani2019density}, we see that most density is concentrated near 0. This suggests that there is a great deal of redundancy in the latent spaces learned by these models, and their true dimensionality is likely significantly less than the full $P$ parameters of the $\theta_{sh}$ weight vector. Unlike these works, the spectral densities are more symmetric around 0. This matches~\cref{fig:trace_figs}, since the Hessian trace (sum of all its eigenvalues) varies between being very positive and very negative. The exact spectral densities vary across properties, even at the end of training.

\section{Discussion}

In this work, we have posited that the performance of multi-task \glspl{GN} for property prediction may be stymied by underlying variation in property-specific loss-surface curvatures. Some existing work considers this question~\cite{Yu2020Surgery} in a simplified theoretical setting, but, to the best of our knowledge, no one has empirically investigated the Hessian properties of multi-task \glspl{GN}. In two domains -- chemistry and materials science -- our results suggest that loss surface curvature varies across each modeled property.

In order to assess curvature without calculating the full Hessian, we build on recent work that uses matrix-free methods to estimate Hessian properties ~\cite{Alain2019negativeeigenvalues,Yao2019pyhessian,Ghorbani2019density,Papyan2020spectra}. We extend their results by considering a novel type of learning problem -- multi-output scalar regression -- and a novel class of neural network -- graph networks. Our results echo some previous results -- the majority of a Hessian's spectral density is concentrated near 0 -- and diverges in other respects -- Hessian properties appear far noisier and more variable for \glspl{GN} than for \gls{CV} tasks. We leave further investigation of this question to future work. Potential explanations include the difference in loss functions for regression vs. classification tasks, a higher level of noise in the datasets we examined, and some special characteristic of \gls{GN} vs. other neural network architectures.

We have here focused on a specific but representative subset of \gls{GN} prediction problems, but considerable potential variation exists. For example, many recent \gls{GN} architectures incorporate a notion of equivariance~\cite{Sattoras2021equivariant,Gasteiger2020dimenet} into their feature-extraction models, and this might impact their curvature in different ways. In addition, we have not evaluated how the choice of optimizer (e.g., stochastic gradient descent vs. Adam~\cite{Kingma2014adam} vs. AdamW) impacts the curvature properties of a learned loss surface.

Our curvature assessment enable several future research directions. First, our analysis here was primarily empirical. The phenomena we identify here (a diversity of curvatures across multi-task loss functions) and the previously-observed degradation of performance for multi-task models~\cite{Gasteiger2020dimenet} could be connected by a theoretical justification. Similarly, we find that curvature properties of \glspl{GN} appear to be much noisier and more variable than the properties of other network architectures, and we currently lack a theoretical justification of why. Intermediate steps might entail investigating the curvature properties of \gls{MTL} methods that, in other domains, do out-perform single-task models (e.g.,~\cite{Sener2018Multiobjective}). Existing work in analyzing curvature for computational geometry (e.g.~\cite{Goldman2005Curvature}) might provide techniques to build upon.

\section*{Acknowledgements}

This work was supported by internal research and development funding from the Research and Exploratory Development Mission Area of the Johns Hopkins University Applied Physics Laboratory. Thanks to Christopher Ratto for assistance in editing and improving the work.




\bibliography{references}

\begin{thebibliography}{35}
\providecommand{\natexlab}[1]{#1}
\providecommand{\url}[1]{\texttt{#1}}
\expandafter\ifx\csname urlstyle\endcsname\relax
  \providecommand{\doi}[1]{doi: #1}\else
  \providecommand{\doi}{doi: \begingroup \urlstyle{rm}\Url}\fi

\bibitem[Alain et~al.(2019)Alain, Roux, and
  Manzagol]{Alain2019negativeeigenvalues}
Alain, G., Roux, N.~L., and Manzagol, P.-A.
\newblock Negative eigenvalues of the hessian in deep neural networks, 2019.
\newblock URL \url{https://arxiv.org/abs/1902.02366}.

\bibitem[Avron \& Toledo(2011)Avron and Toledo]{Avron2011trace}
Avron, H. and Toledo, S.
\newblock Randomized algorithms for estimating the trace of an implicit
  symmetric positive semi-definite matrix.
\newblock \emph{J. ACM}, 2011.
\newblock \doi{10.1145/1944345.1944349}.

\bibitem[Ba et~al.(2016)Ba, Kiros, and Hinton]{Ba2016LayerNorm}
Ba, J.~L., Kiros, J.~R., and Hinton, G.~E.
\newblock Layer normalization, 2016.
\newblock URL \url{https://arxiv.org/abs/1607.06450}.

\bibitem[Battaglia et~al.(2018)Battaglia, Hamrick, Bapst, Sanchez-Gonzalez,
  Zambaldi, et~al.]{Battaglia2018graphnetworks}
Battaglia, P.~W., Hamrick, J.~B., Bapst, V., Sanchez-Gonzalez, A., Zambaldi,
  V., et~al.
\newblock Relational inductive biases, deep learning, and graph networks, 2018.
\newblock URL \url{https://arxiv.org/abs/1806.01261}.

\bibitem[Caruana(1997)]{Caruana1997mtl}
Caruana, R.
\newblock Multitask learning.
\newblock \emph{Machine Learning}, 1997.
\newblock \doi{10.1023/A:1007379606734}.

\bibitem[Chen et~al.(2019)Chen, Ye, Zuo, Zheng, and Ong]{chen2019megnet}
Chen, C., Ye, W., Zuo, Y., Zheng, C., and Ong, S.~P.
\newblock Graph networks as a universal machine learning framework for
  molecules and crystals.
\newblock \emph{Chemistry of Materials}, 2019.
\newblock \doi{10.1021/acs.chemmater.9b01294}.

\bibitem[Gasteiger et~al.(2020)Gasteiger, Groß, and
  Günnemann]{Gasteiger2020dimenet}
Gasteiger, J., Groß, J., and Günnemann, S.
\newblock Directional message passing for molecular graphs.
\newblock In \emph{International Conference on Learning Representations}, 2020.

\bibitem[Ghorbani et~al.(2019)Ghorbani, Krishnan, and
  Xiao]{Ghorbani2019density}
Ghorbani, B., Krishnan, S., and Xiao, Y.
\newblock An investigation into neural net optimization via hessian eigenvalue
  density.
\newblock In \emph{Proceedings of the 36th International Conference on Machine
  Learning}, pp.\  2232--2241, 09--15 Jun 2019.

\bibitem[Gilmer et~al.(2017)Gilmer, Schoenholz, Riley, Vinyals, and
  Dahl]{Gilmer2017Neural}
Gilmer, J., Schoenholz, S.~S., Riley, P.~F., Vinyals, O., and Dahl, G.~E.
\newblock Neural message passing for quantum chemistry.
\newblock In \emph{Proc. of the 34th Int. Conf. on Machine Learning - Volume
  70}, 2017.

\bibitem[Godwin et~al.(2022)Godwin, Schaarschmidt, Gaunt, Sanchez-Gonzalez,
  Rubanova, et~al.]{Godwin2022simple}
Godwin, J., Schaarschmidt, M., Gaunt, A.~L., Sanchez-Gonzalez, A., Rubanova,
  Y., et~al.
\newblock Simple {GNN} regularisation for 3d molecular property prediction and
  beyond.
\newblock In \emph{Int. Conf. on Learning Representations}, 2022.

\bibitem[Goldman(2005)]{Goldman2005Curvature}
Goldman, R.
\newblock Curvature formulas for implicit curves and surfaces.
\newblock \emph{Computer Aided Geometric Design}, 2005.
\newblock \doi{10.1016/j.cagd.2005.06.005}.
\newblock Geometric Modelling and Differential Geometry.

\bibitem[Golub \& Welsch(1969)Golub and Welsch]{Golub1969gauss}
Golub, G.~H. and Welsch, J.~H.
\newblock Calculation of gauss quadrature rules.
\newblock \emph{Mathematics of Computation}, pp.\  221--s10, 1969.

\bibitem[Hill(1952)]{Hill1952vrh}
Hill, R.
\newblock The elastic behaviour of a crystalline aggregate.
\newblock \emph{Proc. of the Phys. Soc. Section A}, 65\penalty0 (5):\penalty0
  349--354, may 1952.
\newblock \doi{10.1088/0370-1298/65/5/307}.

\bibitem[Hättig \& Weigend(2000)Hättig and Weigend]{Haettig2000cc2}
Hättig, C. and Weigend, F.
\newblock Cc2 excitation energy calculations on large molecules using the
  resolution of the identity approximation.
\newblock \emph{J. Chem. Phys.}, 2000.
\newblock \doi{10.1063/1.1290013}.

\bibitem[Jain et~al.(2013)Jain, Ong, Hautier, Chen, Richards,
  et~al.]{Jain2013MaterialsProject}
Jain, A., Ong, S.~P., Hautier, G., Chen, W., Richards, W.~D., et~al.
\newblock {The Materials Project: A materials genome approach to accelerating
  materials innovation}.
\newblock \emph{APL Materials}, 2013.
\newblock \doi{10.1063/1.4812323}.

\bibitem[Kingma \& Ba(2014)Kingma and Ba]{Kingma2014adam}
Kingma, D.~P. and Ba, J.
\newblock Adam: A method for stochastic optimization, 2014.
\newblock URL \url{https://arxiv.org/abs/1412.6980}.

\bibitem[Kong et~al.(2021)Kong, Guevarra, Gomes, and
  Gregoire]{Kong2021multitask}
Kong, S., Guevarra, D., Gomes, C.~P., and Gregoire, J.~M.
\newblock Materials representation and transfer learning for multi-property
  prediction.
\newblock \emph{Appl. Phys. Rev.}, 2021.
\newblock \doi{10.1063/5.0047066}.

\bibitem[Lin et~al.(2016)Lin, Saad, and Yang]{Lin2016lanczos}
Lin, L., Saad, Y., and Yang, C.
\newblock Approximating spectral densities of large matrices.
\newblock \emph{SIAM Review}, 2016.
\newblock \doi{10.1137/130934283}.

\bibitem[Liu \& Nocedal(1989)Liu and Nocedal]{Liu1989lbfgs}
Liu, D.~C. and Nocedal, J.
\newblock On the limited memory bfgs method for large scale optimization.
\newblock \emph{Mathematical Programming}, 1989.
\newblock \doi{10.1007/BF01589116}.

\bibitem[Loshchilov \& Hutter(2019)Loshchilov and Hutter]{Loshchilov2018adamw}
Loshchilov, I. and Hutter, F.
\newblock Decoupled weight decay regularization.
\newblock In \emph{Int. Conf. on Learning Representations}, 2019.

\bibitem[Papyan(2020)]{Papyan2020spectra}
Papyan, V.
\newblock Traces of class/cross-class structure pervade deep learning spectra.
\newblock \emph{Journal of Machine Learning Research}, 21\penalty0
  (252):\penalty0 1--64, 2020.

\bibitem[Park \& Wolverton(2020)Park and Wolverton]{Park2020icgcnn}
Park, C.~W. and Wolverton, C.
\newblock Developing an improved crystal graph convolutional neural network
  framework for accelerated materials discovery.
\newblock \emph{Phys. Rev. Materials}, 2020.
\newblock \doi{10.1103/PhysRevMaterials.4.063801}.

\bibitem[Pearlmutter(1994)]{Pearlmutter1994hessian}
Pearlmutter, B.~A.
\newblock {Fast Exact Multiplication by the Hessian}.
\newblock \emph{Neural Computation}, 1994.
\newblock \doi{10.1162/neco.1994.6.1.147}.

\bibitem[Ramakrishnan et~al.(2015)Ramakrishnan, Hartmann, Tapavicza, and von
  Lilienfeld]{Ramakrishnan2015QM8}
Ramakrishnan, R., Hartmann, M., Tapavicza, E., and von Lilienfeld, O.~A.
\newblock Electronic spectra from tddft and machine learning in chemical space.
\newblock \emph{The Journal of Chemical Physics}, 2015.
\newblock \doi{10.1063/1.4928757}.

\bibitem[Sagun et~al.(2017)Sagun, Evci, Guney, Dauphin, and
  Bottou]{Sagun2017hessian}
Sagun, L., Evci, U., Guney, V.~U., Dauphin, Y., and Bottou, L.
\newblock Empirical analysis of the hessian of over-parametrized neural
  networks, 2017.
\newblock URL \url{https://arxiv.org/abs/1706.04454}.

\bibitem[Sanyal et~al.(2018)Sanyal, Balachandran, Yadati, Kumar, Rajagopalan,
  Sanyal, and Talukdar]{Sanyal2018MTCGCNN}
Sanyal, S., Balachandran, J., Yadati, N., Kumar, A., Rajagopalan, P., Sanyal,
  S., and Talukdar, P.
\newblock Mt-cgcnn: Integrating crystal graph convolutional neural network with
  multitask learning for material property prediction, 2018.
\newblock URL \url{https://arxiv.org/abs/1811.05660}.

\bibitem[Satorras et~al.(2021)Satorras, Hoogeboom, and
  Welling]{Sattoras2021equivariant}
Satorras, V.~G., Hoogeboom, E., and Welling, M.
\newblock E(n) equivariant graph neural networks.
\newblock In \emph{Proceedings of the 38th International Conference on Machine
  Learning}, pp.\  9323--9332, 18--24 Jul 2021.

\bibitem[Sener \& Koltun(2018)Sener and Koltun]{Sener2018Multiobjective}
Sener, O. and Koltun, V.
\newblock Multi-task learning as multi-objective optimization.
\newblock In \emph{Proc. 32nd Int. Conf. on Neural Information Processing
  Systems}, NIPS'18, 2018.

\bibitem[Tan et~al.(2021)Tan, Li, Shi, and Yang]{Tan2021multitask}
Tan, Z., Li, Y., Shi, W., and Yang, S.
\newblock A multitask approach to learn molecular properties.
\newblock \emph{J. Chem. Inf. Mod.}, 61\penalty0 (8):\penalty0 3824--3834,
  2021.
\newblock \doi{10.1021/acs.jcim.1c00646}.
\newblock PMID: 34289687.

\bibitem[Weininger(1988)]{Weininger1988smiles}
Weininger, D.
\newblock Smiles, a chemical language and information system. 1. introduction
  to methodology and encoding rules.
\newblock \emph{J. Chem. Inf. Comp. Sci.}, 1988.
\newblock \doi{10.1021/ci00057a005}.

\bibitem[Wu et~al.(2018)Wu, Ramsundar, Feinberg, Gomes, Geniesse,
  et~al.]{Wu2018moleculenet}
Wu, Z., Ramsundar, B., Feinberg, E.~N., Gomes, J., Geniesse, C., et~al.
\newblock Moleculenet: a benchmark for molecular machine learning.
\newblock \emph{Chem. Sci.}, 9:\penalty0 513--530, 2018.
\newblock \doi{10.1039/C7SC02664A}.

\bibitem[Xie \& Grossman(2018)Xie and Grossman]{Xie2018CGCNN}
Xie, T. and Grossman, J.~C.
\newblock Crystal graph convolutional neural networks for an accurate and
  interpretable prediction of material properties.
\newblock \emph{Phys. Rev. Lett.}, 2018.
\newblock \doi{10.1103/PhysRevLett.120.145301}.

\bibitem[Yang et~al.(2021)Yang, Su, Wang, Jin, Ren, et~al.]{Yang2021multitask}
Yang, A., Su, Y., Wang, Z., Jin, S., Ren, J., et~al.
\newblock A multi-task deep learning neural network for predicting
  flammability-related properties from molecular structures.
\newblock \emph{Green Chem.}, 2021.
\newblock \doi{10.1039/D1GC00331C}.

\bibitem[Yao et~al.(2019)Yao, Gholami, Keutzer, and Mahoney]{Yao2019pyhessian}
Yao, Z., Gholami, A., Keutzer, K., and Mahoney, M.
\newblock Pyhessian: Neural networks through the lens of the hessian, 2019.
\newblock URL \url{https://arxiv.org/abs/1912.07145}.

\bibitem[Yu et~al.(2020)Yu, Kumar, Gupta, Levine, Hausman, and
  Finn]{Yu2020Surgery}
Yu, T., Kumar, S., Gupta, A., Levine, S., Hausman, K., and Finn, C.
\newblock Gradient surgery for multi-task learning.
\newblock In \emph{Adv. Neural Inf. Proc Sys.}, 2020.

\end{thebibliography}
\bibliographystyle{icml2022}

\newpage
\appendix
\onecolumn





\section{Model specifics}\label{app:model}

We use $M=5$ message-passing steps. Node, edge, and global features are projected into a 64-dimensional feature space. $\phi^V$ and $\phi^E$ have two layers with 256 units and a skip connection, followed by an output layer of 64 units. $\phi^U$ has two layers with 192 units and a skip connection, followed by an output layer of 64 units. All networks use $\tanh$ activations.

\section{Training specifics}\label{app:training}

Models are trained for 512 epochs using the AdamW~\cite{Loshchilov2018adamw} optimizer and the default hyperparameters used in the \texttt{optax} implementation. The initial rate is set as $10^{-3}$, with an exponential decay rate of 0.997 applied every epoch after the first 256 epochs.

\section{Data}\label{app:data}

We summarize our datasets in~\cref{table:dataset_stats}.

We scraped Materials Project for crystal records present in it as of October 2020 using the \texttt{MPRester} class from the \texttt{pymatgen} package. The Supplemental Information contains a table of MP IDs used in this study. The raw entries of $G_{VRH}$, $K_{VRH}$, and Poisson ratio $\mu$ contained several anomalously small and large values, so we removed entries with values less than the 5th percentile or greater than the 95th percentile of obtained elastic properties. This left us with a total dataset of 10,500 crystals. Summary statistics for the targets used are given in~\cref{table:mp_stats}. Furthermore, $G_{VRH}$ and $K_{VRH}$ were log-transformed to create a more normal distribution. Roughly 70\% of the dataset (7,400 crystals) was used as training data, and all targets were standardized prior to training.

As initial node features, we used a one-hot encoding of atom type. For initial edge features, we used four features calculated by \texttt{pymatgen}: \texttt{area}, \texttt{face\_dist}, \texttt{solid\_angle}, and \texttt{volume}. Following similar work in applying graph neural networks to crystals~\cite{Xie2018CGCNN}, we discretize these four features based on deciles.

We use the version of QM8 hosted by MoleculeNet~\cite{Wu2018moleculenet}, except that we drop molecules with negative oscillator energies. Summary statistics for the targets are given in~\cref{table:qm8_stats}. As initial node features, we use one-hot encodings of atom type. As initial edge features, we use one-hot encodings of bond type. Roughly 70\% of the dataset (15,300 molecules) was used as training data, and all targets were standardized prior to training.

\begin{table}[h]
    \centering
    \caption{Summary statistics of datasets}
    \begin{tabular}{c||c|c|c|c}
        Dataset     &   \# Data Points  &   \# Targets  &   \# Node Features    &   \# Edge Features\\\hline\hline
        MP          &   10,500          &   3           &   112                 &   40\\\hline
        QM8         &   21,725          &   4           &   9                   &   4
    \end{tabular}
    \label{table:dataset_stats}
\end{table}

\begin{table}[h]
    \centering
    \caption{Summary statistics of \gls{MP} targets}
    \begin{tabular}{c||c|c|c}
              & $\log G_{VRH}$ (log(GPa))   & $\log K_{VRH}$ (log(GPa))     & $\mu$ \\\hline\hline
        mean  & 3.55068348                  & 4.3697952                     & 0.30674381  \\\hline
        std   & 0.73963662                  & 0.67564539                    & 0.06501633  \\\hline
        min   & 0.0                         & 2.63905733                    & 0.18        \\\hline
        max   & 4.82028157                  & 5.45103845                    & 0.5 
    \end{tabular}
    \label{table:mp_stats}
\end{table}

\begin{table}[h]
    \centering
    \caption{Summary statistics of QM8 targets}
    \begin{tabular}{c||c|c|c|c}
          & $E_1$      & $E_2$      & $f_1$      & $f_2$     \\\hline\hline
        mean  & 0.22010506 & 0.24894142 & 0.02289806 & 0.04326408 \\\hline
        std   & 0.04382473 & 0.0345149  & 0.05336499 & 0.07312536 \\\hline
        min   & 0.06956711 & 0.10063996 & 0.0          & 0.0          \\\hline
        max   & 0.51383669 & 0.51384601 & 0.61611219 & 0.56561272
    \end{tabular}
    \label{table:qm8_stats}
\end{table}

\section{Supplemental figures}\label{sec:supp-figs}

\begin{figure*}[h]
    \centering
    \begin{subfigure}{0.55\textwidth}
         \centering
         \includegraphics[width=\linewidth]{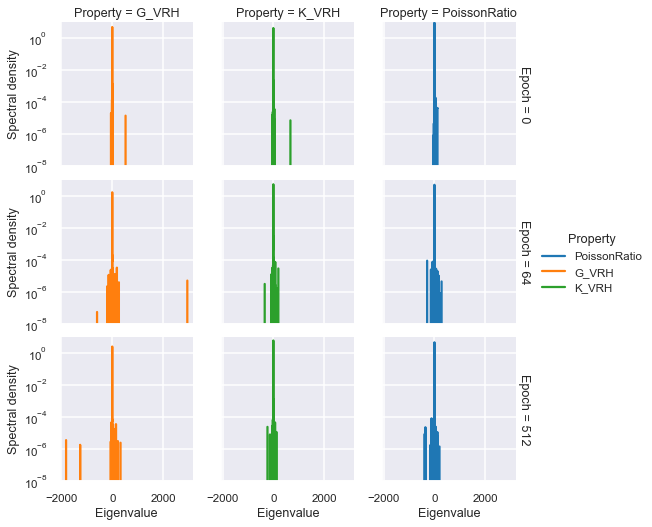}
         \caption{\gls{MP}}
         \label{fig:mp_spectra_full}
     \end{subfigure}
     \hfill
     \begin{subfigure}{0.55\textwidth}
         \centering
         \includegraphics[width=\textwidth]{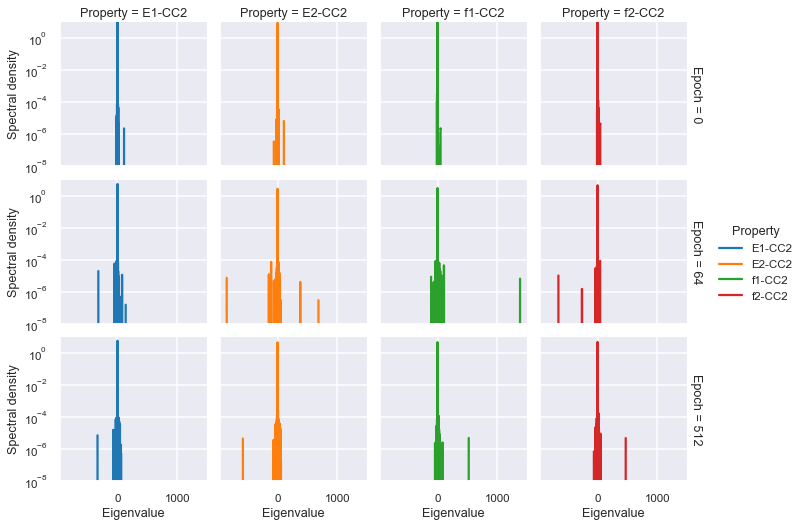}
         \caption{QM8}
         \label{fig:qm8_spectra_full}
     \end{subfigure}
    \caption{We visualize snapshots of property-specific estimated spectral densities during selected training points. For both \gls{MP} and QM8 models, most of the spectral density is concentrated near 0, especially after random initialization. Occasionally, very high-magnitude eigenvalues, both positive and negative, spike up for a short period. Our results here elaborate on~\cref{fig:trace_figs} -- the high and varying curvature of property-specific losses is driven by singularly large eigenvalues that vary across properties. For these plots, the $x$-axis gives the range of eigenvalues for the loss function, and the $y$-axis gives the density $\hat{\psi}(t)$ of eigenvalues concentrated at that point $t$ (as described in~\ref{sec:curvature}).}
    \label{fig:full_spectra}
\end{figure*}

\newpage

\section{Example spectrum estimation}\label{sec:extra-lanczos}

\begin{figure}[h]
    \centering
    \includegraphics[width=0.75\linewidth]{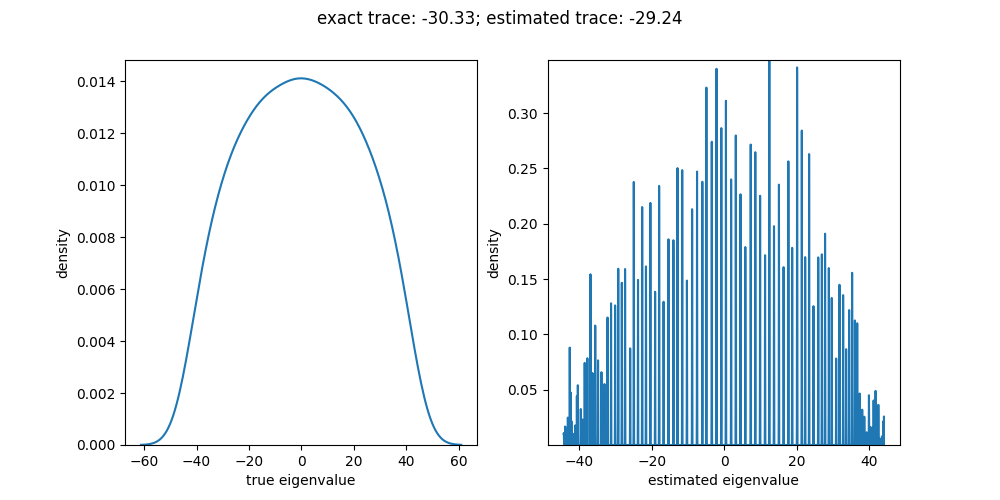}
    \caption{An example showing the use of the Lanczos algorithm to estimate the spectral density of a Hessian. The left figure is the distribution of the true eigenvalues, and the right is the estimated spectral density. The estimated values accurately characterize the maximal and minimal eigenvalues, and the interior shows qualitative agreement with the known distribution.}
    \label{fig:example_lanczos}
\end{figure}

We consider a loss function defined by $L(x) = \dfrac{1}{2}x^T A x$, where $A = \dfrac{1}{2}(B + B^T)$, for a matrix $B$ with entries sampled $iid$ from a standard normal distribution. In this case, the Hessian is given by $A$ and so its estimated eigenvalues can be compared to its true eigenvalues. In~\cref{fig:example_lanczos}, we plot the results of a sample calculation, for a $1,000 \times 1,000$ matrix. We show that the estimated density has reasonable similarity to the true distribution, and the estimated trace also matches the true trace.

\end{document}